\documentclass[a4paper, 10pt, conference]{ieeeconf}      

\IEEEoverridecommandlockouts                              

\usepackage{cite}
\usepackage{balance}
\usepackage{amsmath,amssymb,amsfonts}
\usepackage{soul}
\usepackage{algorithmic}
\usepackage[hidelinks]{hyperref}
\usepackage[linesnumbered,ruled,vlined]{algorithm2e}
\usepackage{graphicx}
\usepackage{textcomp}
\usepackage{xcolor}
\usepackage{caption}
\usepackage{subcaption}
\usepackage{multirow}
\usepackage{balance}

\usepackage{array}
\newcommand{\PreserveBackslash}[1]{\let\temp=\\#1\let\\=\temp}
\newcolumntype{C}[1]{>{\PreserveBackslash\centering}p{#1}}

\title{\LARGE \bf
Object Detection in Thermal Images Using Deep Learning for Unmanned Aerial Vehicles
}

\author{Minh Dang Tu$^{1}$, Kieu Trang Le$^{1}$, Manh Duong Phung$^{2}$
\thanks{$^{1}$Minh Dang Tu, Kieu Trang Le are with Vietnam National University, Hanoi, Vietnam.}
\thanks{$^{2}$Manh Duong Phung is with Fulbright University Vietnam, Ho Chi Minh City, Vietnam. {\tt\footnotesize duong.phung@fulbright.edu.vn}}
}

\begin{document}

\maketitle
\thispagestyle{empty}
\pagestyle{empty}

\begin{abstract}
This work presents a neural network model capable of recognizing small and tiny objects in thermal images collected by unmanned aerial vehicles. Our model consists of three parts, the backbone, the neck, and the prediction head. The backbone is developed based on the structure of YOLOv5 combined with the use of a transformer encoder at the end. The neck includes a BI-FPN block combined with the use of a sliding window and a transformer to increase the information fed into the prediction head. The prediction head carries out the detection by evaluating feature maps with the Sigmoid function. The use of transformers with attention and sliding windows increases recognition accuracy while keeping the model at a reasonable number of parameters and computation requirements for embedded systems. Experiments conducted on public dataset VEDAI and our collected datasets show that our model has a higher accuracy than state-of-the-art methods such as ResNet, Faster RCNN, ComNet, ViT, YOLOv5, SMPNet, and DPNetV3. Experiments on the embedded computer Jetson AGX  show that our model achieves a real-time computation speed with a stability rate of over 90\%.
\end{abstract}

\section{Introduction}
Unmanned aerial vehicles (UAVs) are being used for specialized tasks such as search and rescue, surveillance, and military operations. To carry out these tasks, UAVs are equipped with image sensors and onboard processing units to collect data and then analyze it. The analysis often requires object recognition and classification from which relevant operations can be conducted. For UAV object recognition, color images are commonly used as input data. However, in many environmental conditions such as nighttime or low light, the use of color images is ineffective as objects and the environment blend together. Using thermal images is then considered a more relevant solution with advantages such as the ability to identify small and tiny objects as long as their heat emission is different from the ambient temperature, and the possibility to search in challenging conditions such as fog and low light. The main challenge, however, is how to deal with low resolution, uneven thermal background, and high noise in thermal images. A number of studies have been proposed to address these problems with details as follows.
 
\subsection{Small object detection}
Methods for detecting small objects have been developed for years with state-of-the-art methods such as Viola Jones Detectors \cite{990517}, Histogram of Oriented Gradients (HOG) detector \cite{dalal2005histograms}, Deformable Part-based Model (DPM) \cite{4587597}, etc. However, the development of machine learning techniques, especially deep learning, has made it a trend due to the superior results and advantages over traditional techniques. In deep learning, object detection methods are divided into two main methods, one-stage and two-stage detection. The one-stage method classifies objects and generates bounding boxes without using pre-generated region proposals. It thus has a fast execution time with representatives such as Retical net, SSD\cite{ssd}, Yolo\cite{yolo},  CornerNet, FCOS\cite{fcos}, FSAF\cite{fsaf}. The two-stage method first generates region proposals and then classifies objects in each region proposal. It is therefore more accurate but slower with notable representatives such as mask R-CNN\cite{mask}, Faster R-CNN\cite{frcnn}. Continuous advancements in deep learning have led to the popularity of CNN models for object detection, leading to the development of new models such as VGG, Googlenet, DenseNet, Resnet, Efficientnet\cite{efficientnet}, ViT\cite{vit}, which have higher quality feature maps and can detect objects more accurately.

The progress of object detection models also comes from changing the computation method in individual layers with large computational depth \cite{sspnet},  model ensemble strategies \cite{1st}, and region-based convolutional neural network \cite{1st}... The improvements also include changing recognition layers from global to local feature maps \cite{1st}, using new detection kernels \cite{sspnet}, and combining feature maps \cite{roi}. The accuracy of the models thus gradually increases. However, there are two challenges when using these models: (1) their deployment on an embedded device is relatively complex, and (2) the performance of the models is not yet truly stable.

\subsection{Combining models in object detection}
Deep learning is a flexible and scalable approach that can adapt to the increasing amount of data used for training. One drawback of this method is that it depends on the data used, and when encountering entirely new data, the weights of the network will change accordingly leading to an increase in the variance. One approach to mitigate this is to train multiple models and then combine their predictions.

There are three commonly used methods to combine predicted bounding boxes including the non-maximum suppression (NMS) \cite{nms}, Soft-NMS \cite{soft}, and weighted boxes fusion (WBF) \cite{wbf}. In NMS, if the bounding box overlaps and exceeds a certain Intersection over Union (IoU) threshold, they are considered to belong to one object. NMS then only retains the box with the highest confidence and removes the rest. Soft-NMS improves the NMS by using a function to reduce the confidence of adjacent bounding boxes based on the IoU threshold. WBF operates differently from NMS by merging boxes instead of removing them to produce the final result. For color images provided by UAVs, WBF yields good results. However, with thermal images, this method may mistakenly detect a background area as an object when its thermal dissipation is similar to an object. Therefore in our approach, Soft-NMS is used to combine predicted bounding boxes.

\subsection{Thermal image processing}
Different from color images with three channels, the information of a thermal image is only equivalent to a grayscale image that represents the response of the capturing area to infrared frequencies. Most approaches therefore use the two-stage method to extract sparse features of harsh environments \cite{2020thermal}; learn features that maximize the information between visible and infrared frequencies \cite{2021sauto}; or detect low-resolution objects from UAVs \cite{2021uav}. The use of two-stage methods however limits the deployment of those models to systems that have limited computational capability.

In this paper, we present a CNN model that solves two issues in recognizing objects from thermal images taken by UAVs. The first issue is the limited information of thermal images when representing objects with varying temperatures such as boats or parked vehicles. The second involves objects that are close together, when taken by a UAV, will likely be recognized as a single object. To solve these problems, our approach focuses on extracting feature maps through a multi-dimensional pyramid and using an attention mechanism to enrich data and expand information regions, as shown in Fig.\ref{fig:my_model}. We use attention layers instead of increasing the depth of CNN to reduce computation cost. In addition, we apply the Bi-FPN method instead of other FPN to obtain a feedforward computation process that targets high-value regions. In the final layer, we use Soft-NMS \cite{soft} to combine predicted bounding boxes of the training models.

Our contributions in this work include:
\begin{itemize}
    \item Analyze the characteristics of small object detection problems and study the methods used for object recognition from both color images and thermal images taken by UAVs to come up with a new network model. The model can identify small objects by using sliding windows to remove regions of low correlation with their surrounding areas and output high-potential regions for identification. The model utilizes ViT and YOLOv5, the network architectures that perform well with color images, and augments them with attention blocks to better identify objects.
    
    \item Modify the backbone architecture to its equivalence but with fewer parameters to better use on UAVs. Furthermore, we also carry out the process of compressing the model's weights to ONNX and TensorRT to increase the execution speed of the model. We also use self-training to reinforce the classification ability of objects and eliminate incorrect recognition features during training.
    \item Our model performs well on thermal images subjected to high noise. The model is designed as a module that can be deployed on embedded systems such as UAVs. The model achieves high accuracy in both public and our own datasets.
    
\end{itemize}
\section{The proposed model}
\begin{figure*}
    \centering
    \includegraphics[width=0.85\textwidth]{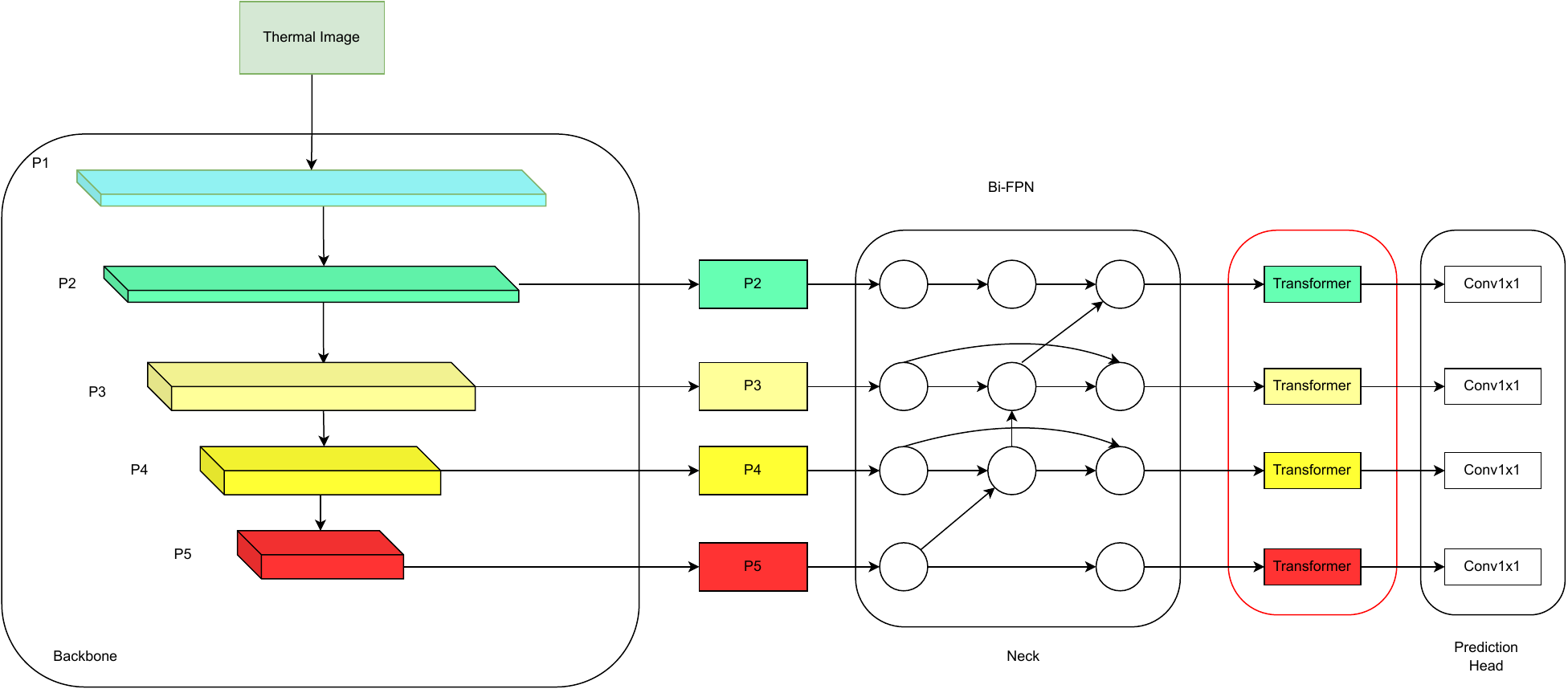}
    \caption{The architecture of our model}
    \label{fig:my_model}
\end{figure*}
In this section, we present the proposed method for object detection using thermal images from UAVs. Our model is built on the architecture of YOLOv5 with the following 3 structures: backbone, neck, and head, as shown in Fig.\ref{fig:my_model}. We use attention layers in both the layers after the backbone and transformers to expand the features of the input matrix to overcome limitations related to the low information of thermal images, especially with small objects, people, and vehicles in the image. The use of attention layers has the role of expanding the computational area and making features of the object clear for identification. Due to the low information content in thermal images, we perform calculations on surrounding pixels to detect the bounding box of the object.

\subsection{Number of prediction heads }\label{AA}
For the YOLOv5 model \cite{yolo}, after passing through the backbone and neck layers, three feature matrices are produced to input into the prediction heads. However, these three prediction heads are only sufficient for cases where the objects in the image occupy a large portion of the image. In our implementation environment, which involves images captured from drones as in the VEDAI \cite{vedai} dataset, objects in the images occupy a small portion of the image. To identify these objects, we use 4 prediction heads instead of 3. The additional prediction head, extracted from the low-level feature layer, helps to reduce variance in identifying the bounding box and increase the ability to detect new objects that might have been omitted in higher feature layers. After adding one prediction head, the number of calculation parameters increases, but the object detection performance is significantly improved.

\subsection{Backbone optimization}\label{AA}
Due to the requirement to use on UAVs, a recognition model that can extract feature maps fast enough is needed. We thus employ the feature extraction methods through the use of GhostConv to reduce the number of parameters that need to be calculated. To address the problem of low thermal information, we use computation layers of Bottleneck net to extract additional information. For an input $X \in R^{c \times h \times w}$, where $c$ is the number of input dimensions, $h$ and $w$ are the size of the input matrix, the value of feature layer $n$ is calculated as:
\begin{align}
  Y = X * f + b,
  \label{ct1}
\end{align}
where $Y \in R^{h'\times w'\times n}$ is the result of feature layer $n$, $b$ is the bias term, and $f \in R^{c \times k \times k \times n}$ is the filter of the convolutional block. The GhostNet method produces the result through the processes of standard convolution and series of linear operations on each intrinsic feature. In standard convolution, $Y \in R^{h'  \times w'  \times m}$ with $m < n$ is obtained from input $X \in R^{c \times h \times w}$ and convolutional block matrices  $f^{'} \in R^{c  \times  k  \times  k  \times m}$ according to formula $Y^{'} = X * f^{'}$. The series of linear operations convert tensor $Y^{'}$ to matrix $Y^{''}$ by the transformation matrix $g(x)$ according to
\begin{align}
y{''}_{ij} = g_{ij}\cdot {y{'}_{ij}} \quad i \in {1, \dots, m} ; j \in {1, \dots, s}
\label{ct2}
\end{align}
Each value of $y{'}$ gives a value of $y{''}$, where $m$ and $s$ are the size of $y{'}$ and $y{''}$. Finally, $y{'}$ and $y{''}$ are combined to generate the output. With the use of $g_{ij}$, the computational cost is lower with fewer parameters than a conventional CNN network.

\subsection{Transformer encoder block}\label{AA}

Inspired by the transformer method applied in Vision Transformer (ViT), we apply the transformer block at the end of the backbone block to replace the Bottleneck net convolution layer in the original version of YOLOv5. The input of this layer is the output of the forward convolution layer and the information distributed through the ASPP block \cite{aspp}. The input of the transformer block is represented as $I_{IR} \in R_{IR}^{c \times h \times w}$ given from feature map $F_{IR} \in R_{IR }^{c \times h \times w}$. $I_{IR}$ is fed into the attention process with 3 parameters $Q$, $K$, $V$ computed as:
\begin{align}
    Q = I_{IR}\cdot W_Q \\
    K = I_{IR}\cdot W_K \\
    V = I_{IR}\cdot W_V, 
\end{align}
where $W_Q$, $W_K$, $W_V$ are matrices trained in the model. The output, $Z$, of the attention block is then calculated as:

\begin{align}
Z = Attention(Q,K,V) = Softmax(\frac{QK^{T}}{\sqrt{d_k}})\cdot V, 
\label{att}
\end{align}
where $d_k$ is the parameter maintaining the convergence of the $Softmax$ result. From $Z$, the activation layers GELU is used to compute the output. Using GELU instead of ReLU allows for performing regression on the data instead of discarding the data with negative parameters, thereby reducing the variance.

\begin{figure*}[t]
\centering
{\includegraphics[width=0.3\textwidth]{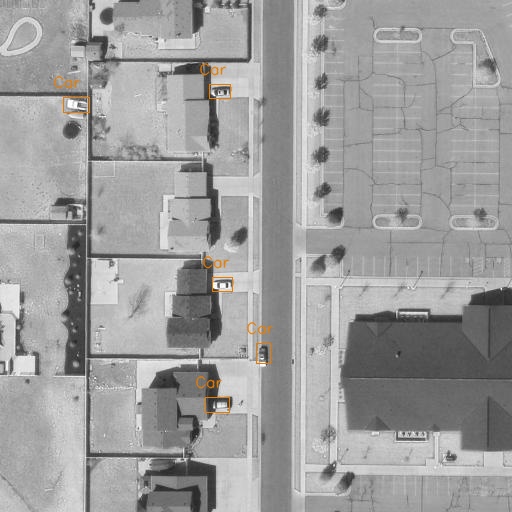}}
{\includegraphics[width=0.3\textwidth]{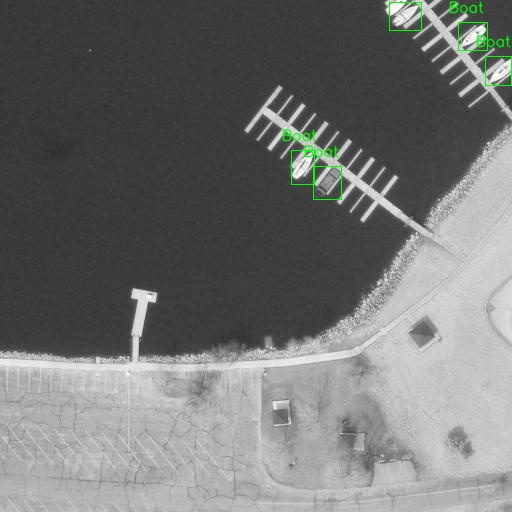}}
{\includegraphics[width=0.3\textwidth]{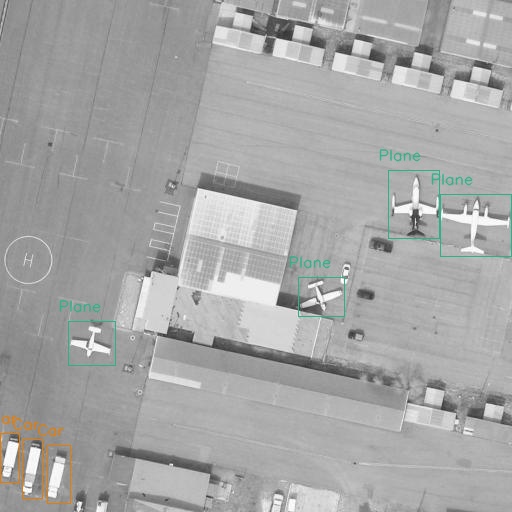}}
\hfill 
\\[1ex]
{\includegraphics[width=0.3\textwidth]{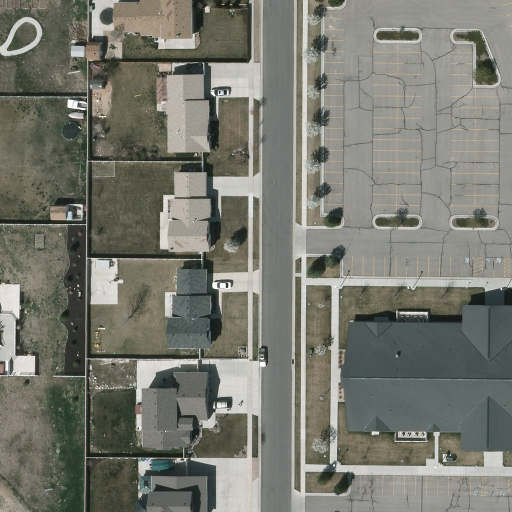}}
{\includegraphics[width=0.3\textwidth]{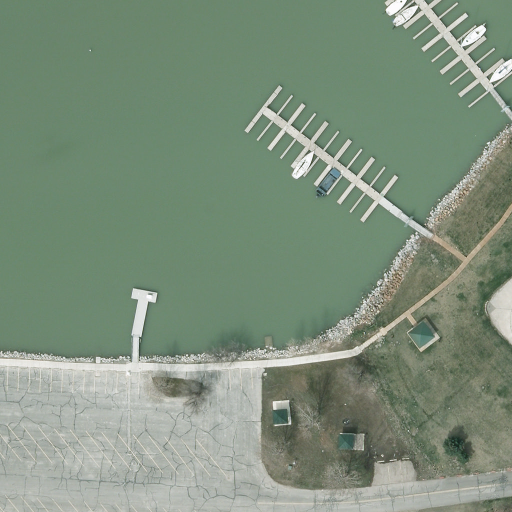}}
{\includegraphics[width=0.3\textwidth]{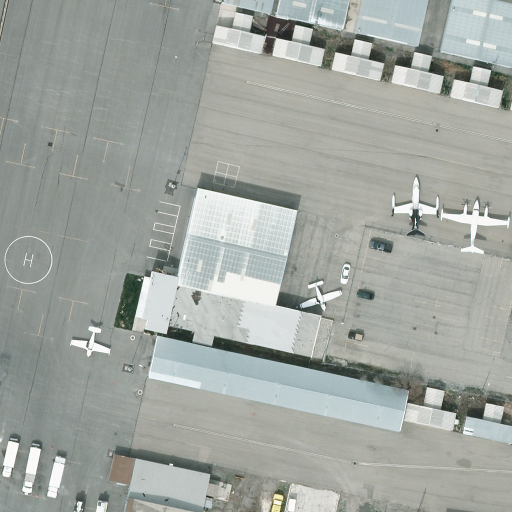}}
\hfill 

\caption{Result in the public dataset VEDAI}
\label{fig:public_data}
\end{figure*}

\begin{figure*}[t]
\centering
\includegraphics[width=0.235\textwidth]{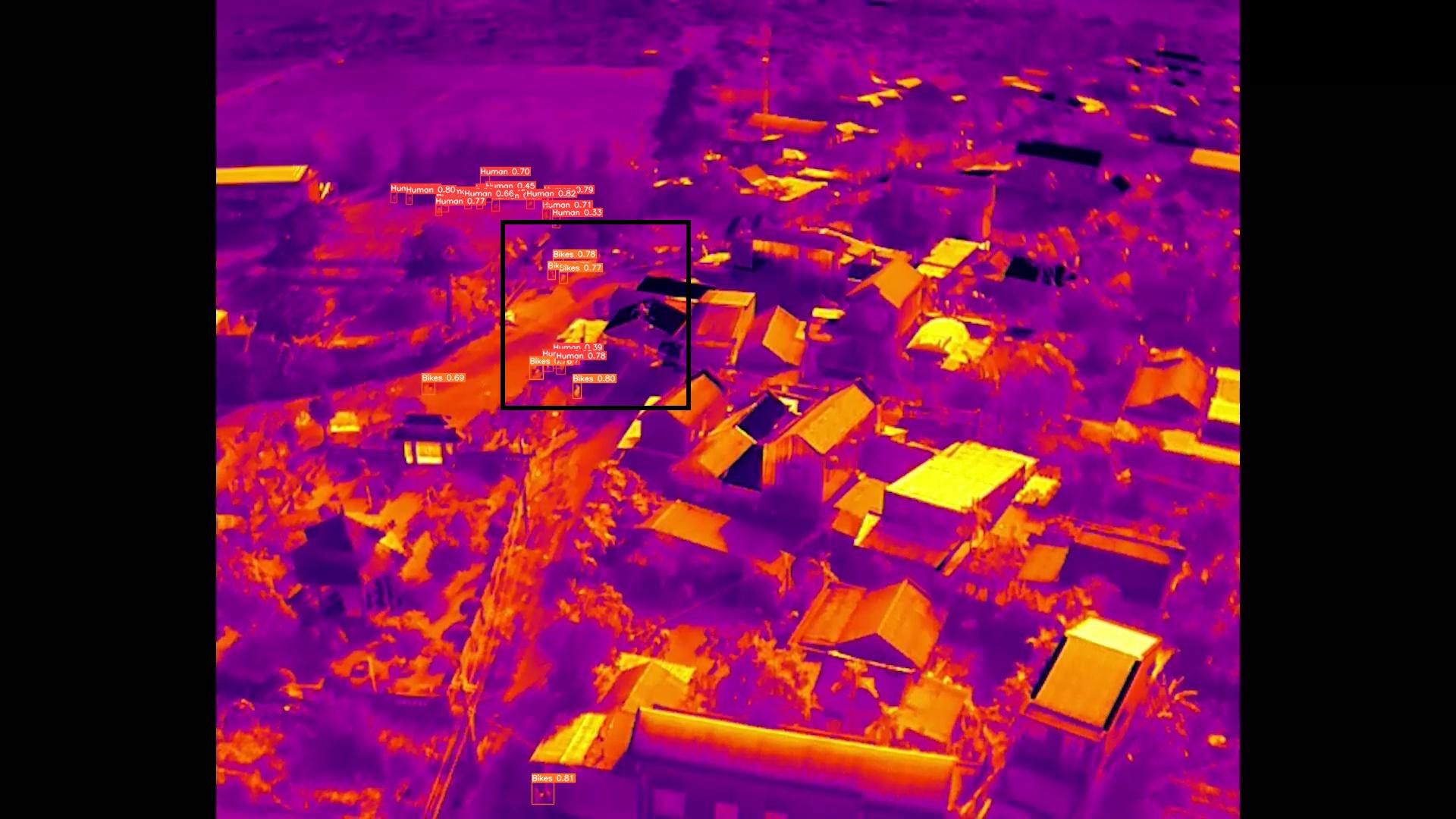}
\hfill 
\includegraphics[width=0.235\textwidth]{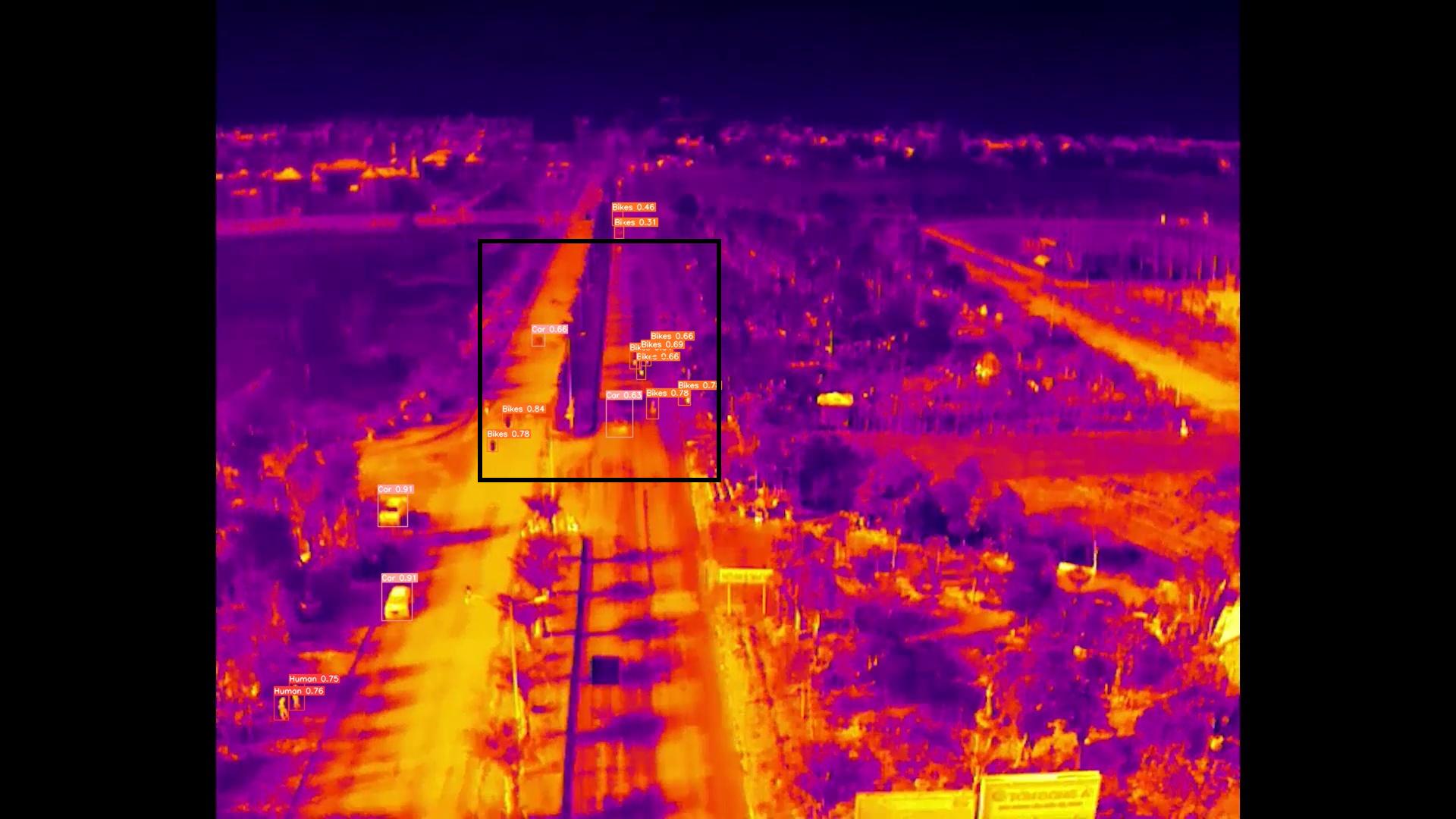}
\hfill 
\includegraphics[width=0.235\textwidth]{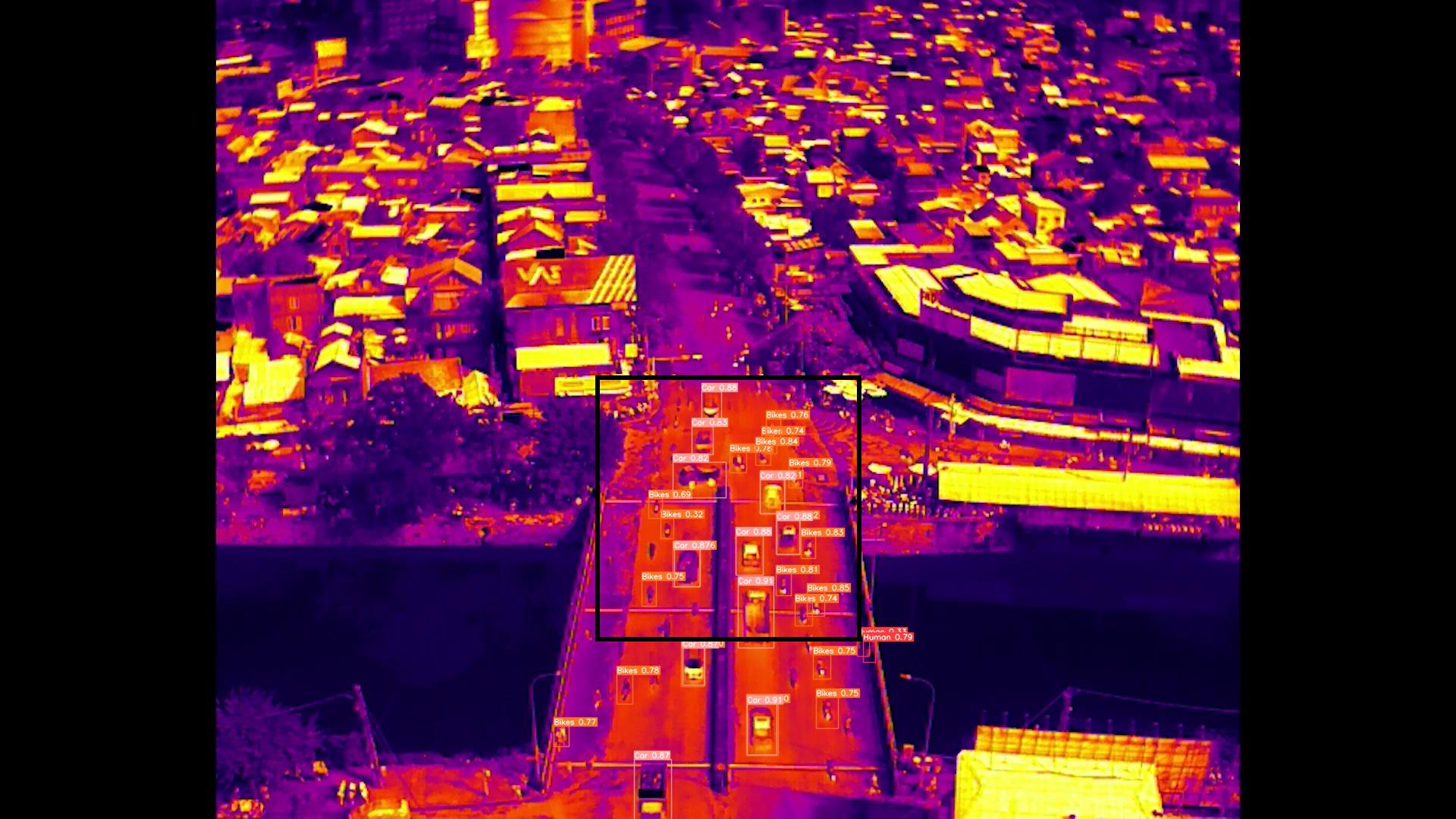}
\hfill 
\includegraphics[width=0.235\textwidth]{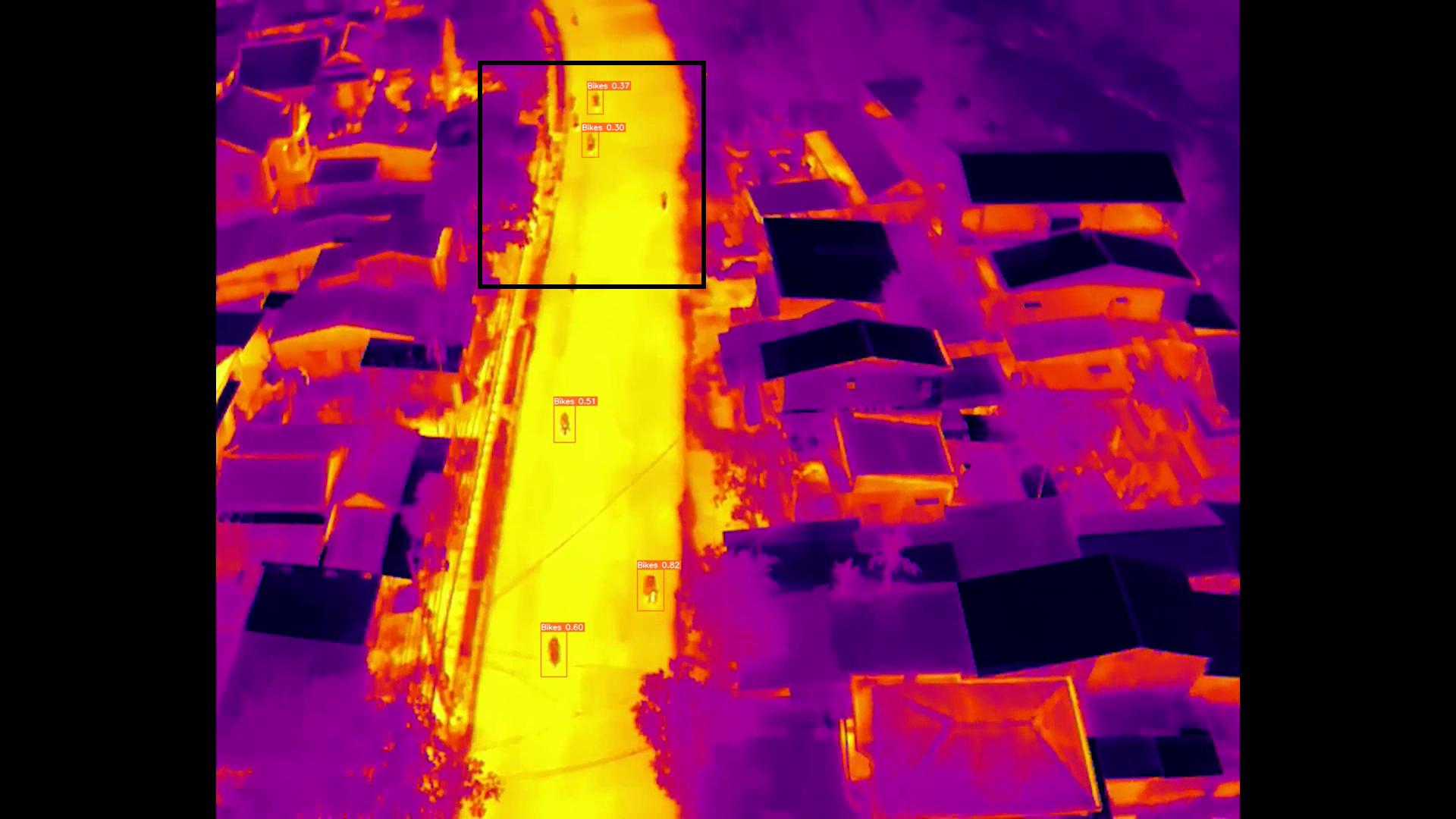}
\\[1ex]
\includegraphics[width=0.235\textwidth]{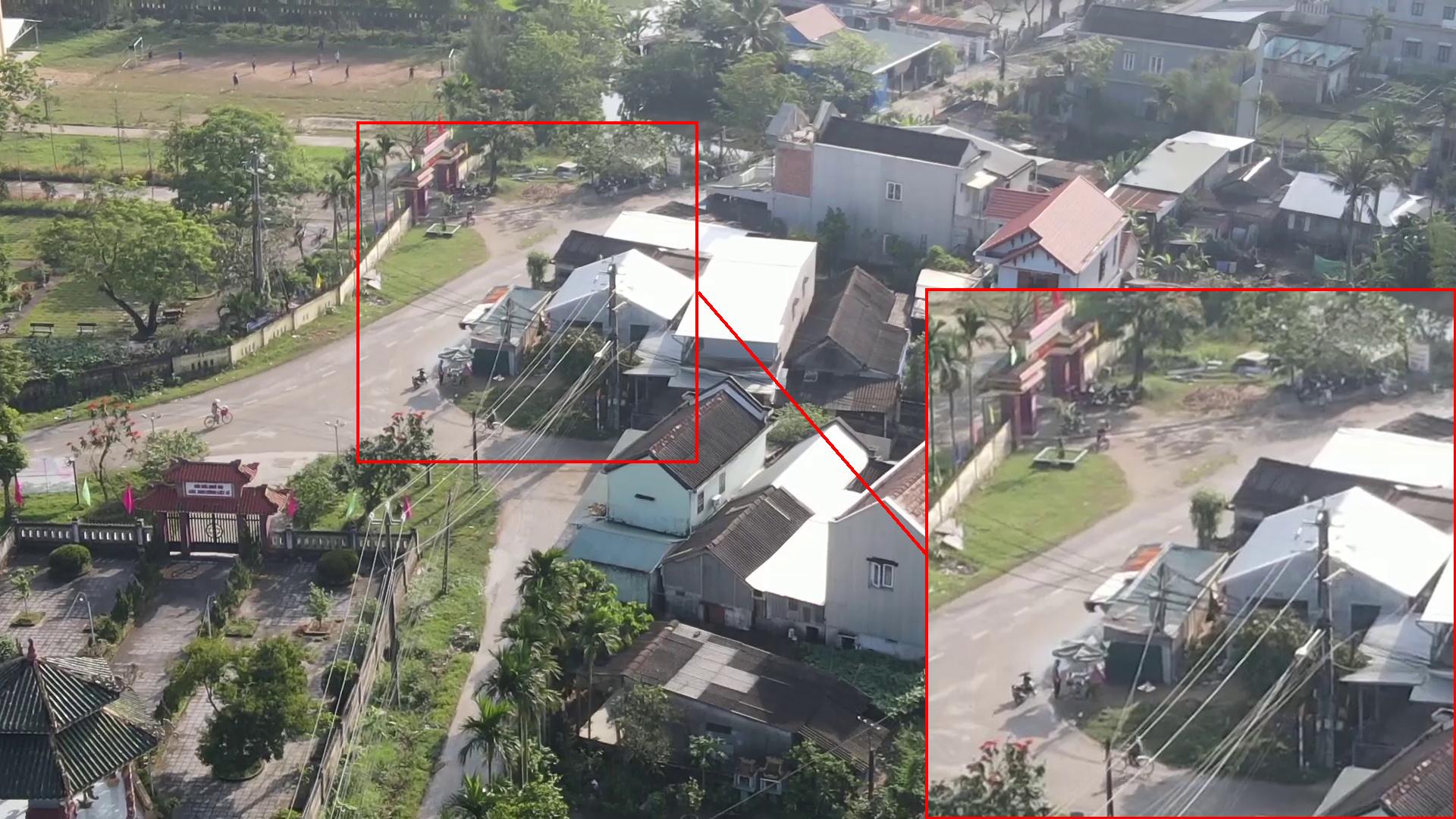}
\hfill 
\includegraphics[width=0.235\textwidth]{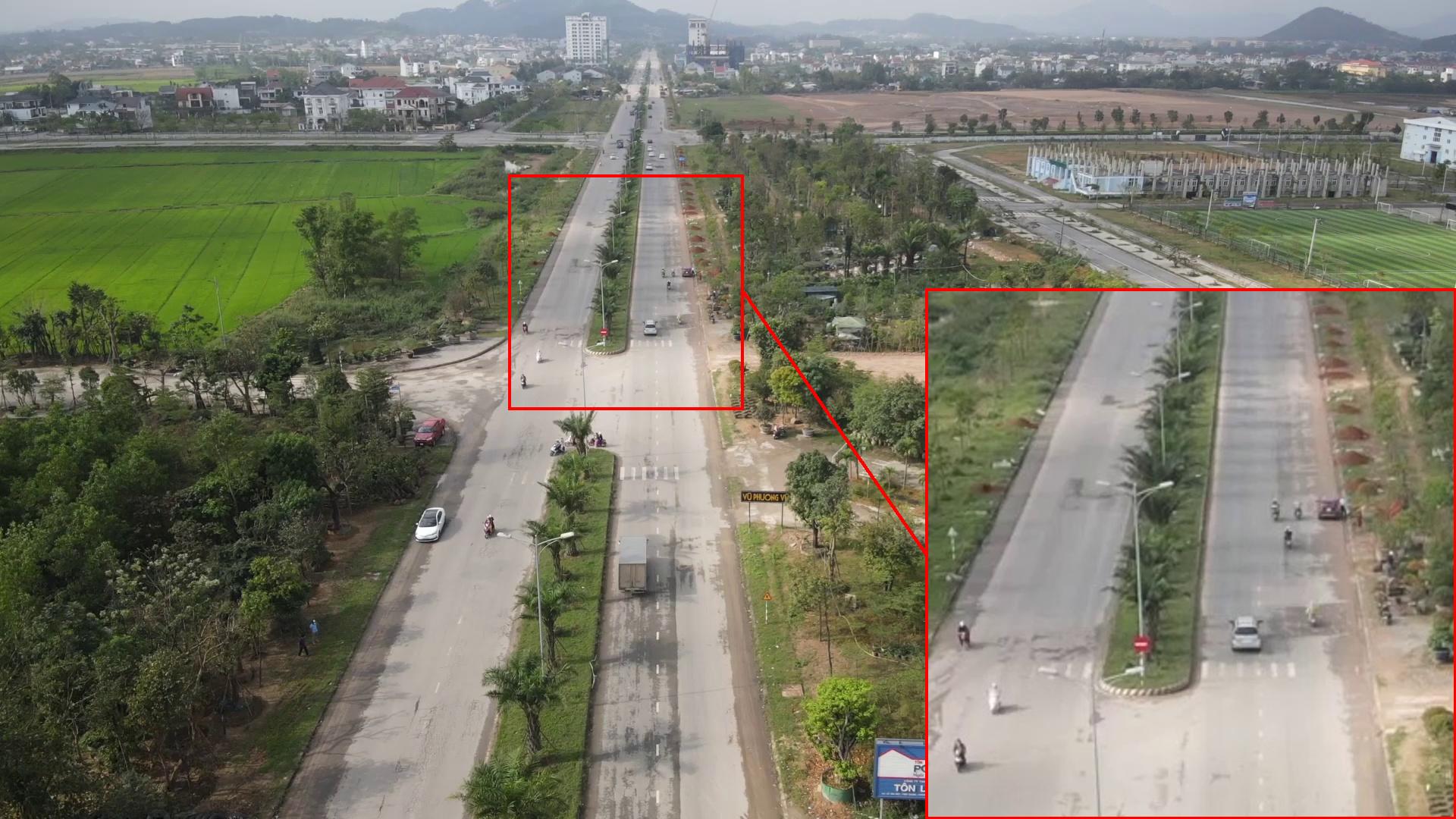}
\hfill 
\includegraphics[width=0.235\textwidth]{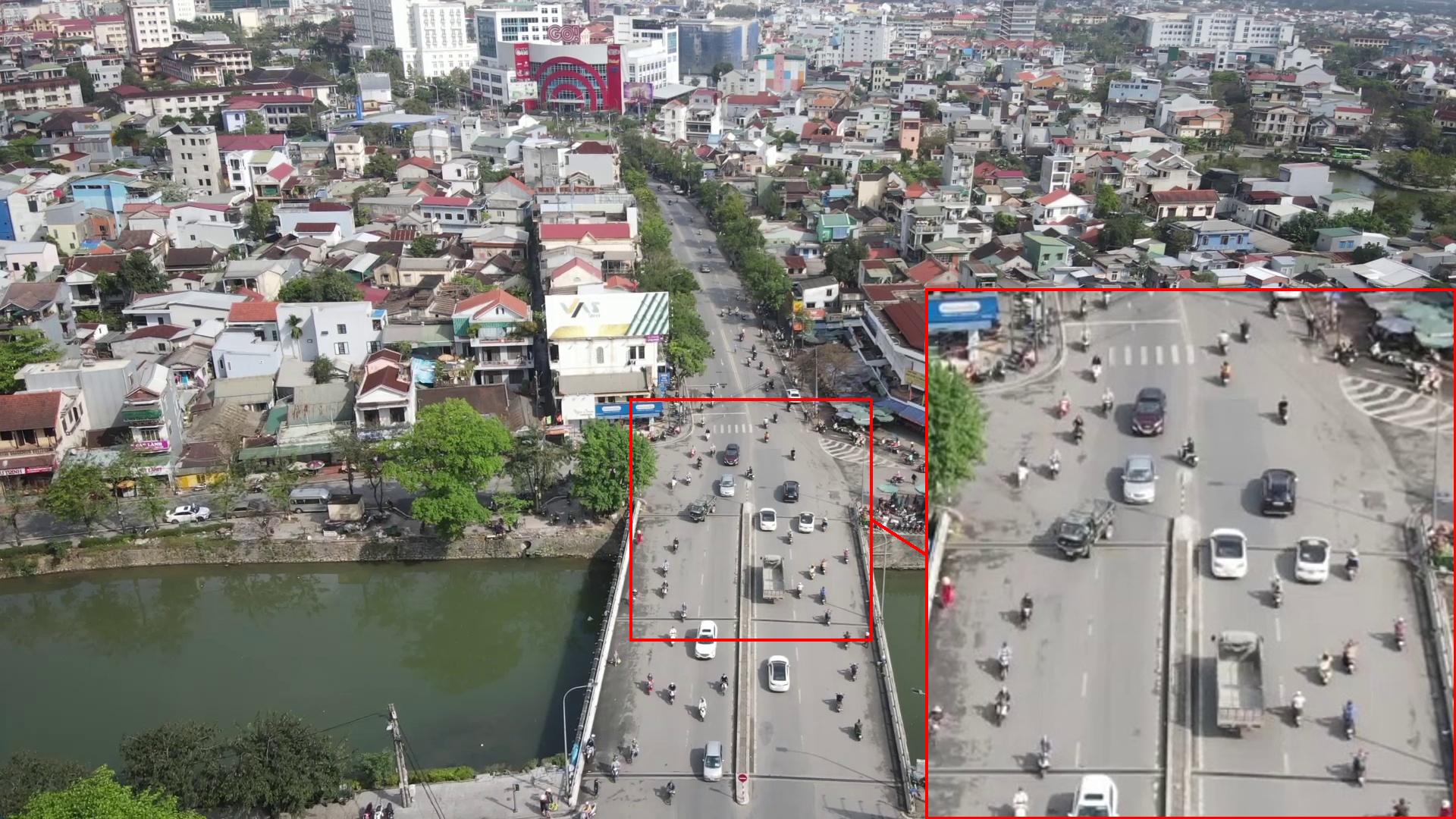}
\hfill 
\includegraphics[width=0.235\textwidth]{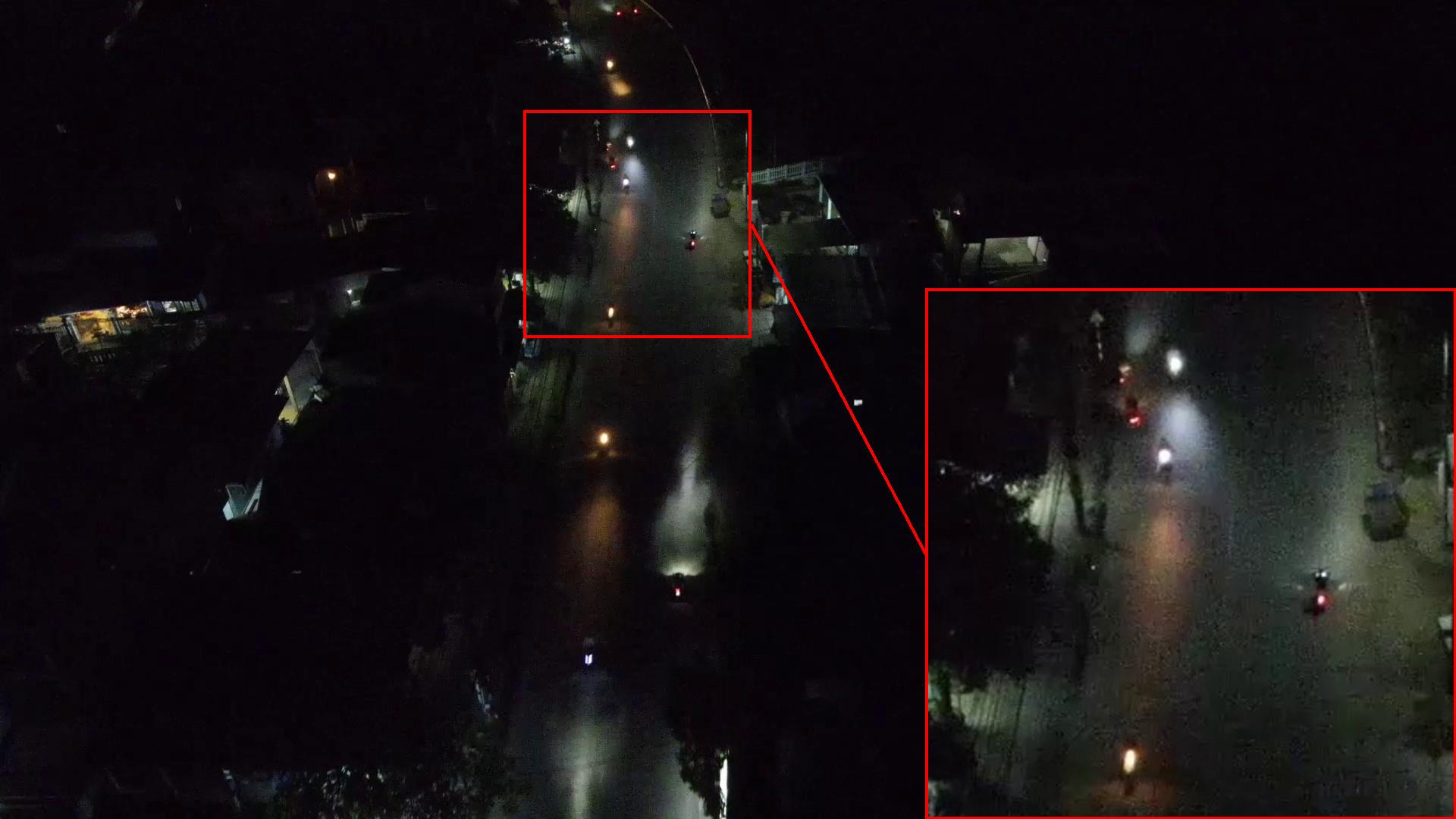}
\\[1ex]
\subcaptionbox{}{\includegraphics[width=0.235\textwidth]{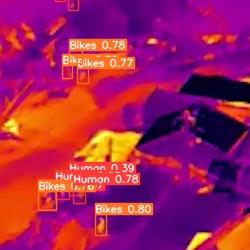}}
\hfill 
\subcaptionbox{}{\includegraphics[width=0.235\textwidth]{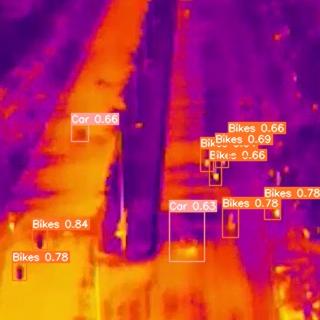}}
\hfill 
\subcaptionbox{}{\includegraphics[width=0.235\textwidth]{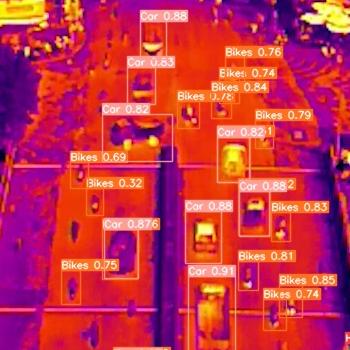}}
\hfill 
\subcaptionbox{}{\includegraphics[width=0.235\textwidth]{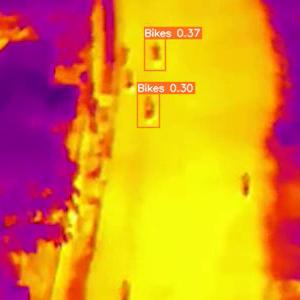}}
\caption{Result with our dataset in different scenarios: (a) and (b) normal scenarios, (c) City with dense vehicles and (d) Night time when the visual images are totally black}
\label{fig:our_data}
\end{figure*}


\subsection{Convolution using sliding windows and attention}
In our model, we combine the use of sliding windows with the attention mechanism instead of using the usual Transformer method to achieve fast object detection. This module is also simple enough to integrate into the CNN system without causing the computation system to exceed the processing amount in the YOLOv5 architecture. This block has two processes including splitting the input matrix by a sliding window and performing Attention on the obtained windows.  

The input of this module is the output feature maps of the Bi-FPN. With the feature map having size $(h \times w \times c)$ through the sliding window mechanism, it is reduced to size $(n \times mh \times mw \times c)$,along with creating a mask matrix of size size $(n \times mh\cdot mw \times mh\cdot mw \times c)$. With the values $h$, $w$, $c$ being the width, height, and depth of the input feature matrix, respectively; $mh$, $mw$ are the length and width of the sliding window respectively.
Self-attention is then computed on the sliding windows and their surroundings to extract features on those windows. Here, we use GELU layers to ensure accuracy when working with large data. The output is computed as:
\begin{align}
    \hat{z}^l = WA(LN(z^{l-1}) +z^{l-1}\\
    z^l = MLP(LN(\hat{z}^l)) +\hat{z}^l,
\end{align}
where $z^l$ and $\hat{z}^l$ are respectively the outputs of the windows attention (WA) and multilayer perception (MLP) modules in block $l$; WA represents a multi-head self-attention block using sliding windows; LN denotes a linear block with linear operations on $z$. The result of the block is the feature map augmented with feature information extracted by the attention layers. This module thus helps to identify the areas of interest that the model should focus on to detect objects.

\subsection{Training and execution process}
For large models, the process of detecting objects can take a lot of time. During training, we adjust the input size to minimize the loss of information with the input image size being 0.67 and 0.85 of the original image size. We also rotate and flip images to diversify the training data. In the experiment, we decided to use the input size of 0.67 of the original image size, and retain the grayscale or color scale of the thermal image. We use techniques such as layer stacking, and expanding the calculation area to increase the recognition speed. We also combine with models having a low number of parameters through the use of ConvGhost and Bottleneck Ghost to reduce the time to compute feature maps.

\section{Results}\label{AA}
To evaluate the proposed model, we have conducted experiments on a public dataset named VEDAI and the datasets collected from our UAV. We have also carried out comparisons with other state-of-the-art models on two platforms including a high-performance server and an embedded computer.

\begin{figure}
    \centering
    \includegraphics[width=0.45\textwidth]{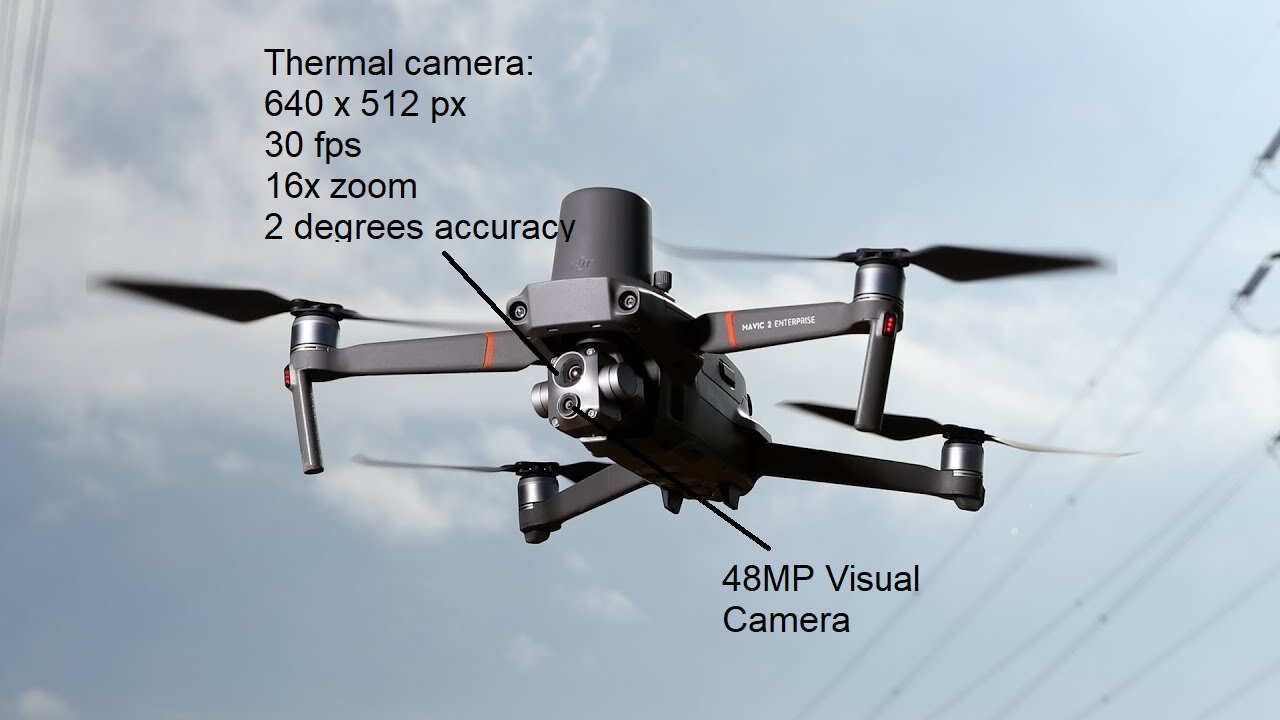}
    \caption{Drone Mavic 2 Enterprise Advanced used to collect data}
    \label{fig:uav}
\end{figure}

\subsection{Datasets}
Our data is collected by the drone named Mavic 2 Enterprise Advanced equipped with cameras as shown in Fig. \ref{fig:uav}. The thermal camera has a resolution of $640 \times 512$ px, speed of 30 fps, 16$\times$ zoom, and accuracy of $\pm2$ Celcius degrees. The drone is also equipped with a color camera having a resolution of 48 MP. This camera, however, only uses for reference purposes. The video data collected has a duration of 11 hours and is recorded at different times and locations. It thus provides us with millions of thermal images for training and testing. Besides, we also use the public dataset named VEDAI to utilize the thermal images collected from the drone in various conditions, especially in low light. Images in the datasets are pre-processed and labeled to ensure their quality when used for the models. In comparisons, all the models use the same image size of $800 \times 800$. 

\subsection{Evaluation metrics and platforms}
In all comparisons, we use the mean average precision (mAP) metric to evaluate the accuracy of the models at two IoU thresholds, 0.5 and 0.95. The platform used for training the models is the servers with GPU A100 3.2G 40G. The optimization function used is Adam with a learning rate of $3.2 \times 10^{-5}$. We use 10 initial epochs for the purpose of stabilizing the training, and 150 actual training epochs with a batch size of 16. This process is performed on both our own datasets and the public dataset.

\subsection{Evaluation results}

\subsubsection{Results}
We compare our model with the popular models used for object detection including Faster R-CNN, ResNet, YOLO net, Vision Transformer (ViT), E-net, ComNet\cite{comnet}, SMPNet \cite{smp}, DPNetV3 \cite{teamvisdrone}. The results are presented in Tables \ref{table:1} and \ref{table:2}. It can be seen that our model is more accurate than other methods at both values 0.5 and 0.95. Compared to methods that use equivalent amounts of parameters, our model shows better results on both datasets. Specifically, our model performs better by 4.1\% to 7.01\% on the VEDAI dataset for mAP50 and by 4.42\% to 9.83\% on our dataset for mAP95. Besides, our model has relatively low number of parameters thanks to the replacement of multiple CNN blocks with computationally efficient layers such as attention layers and feedforward linear layers. As a result, our model achieves higher performance compared to other models and has the ability to be deployed on embedded systems. 
\begin{table}[h!]
\centering
\begin{tabular}{|c|c|c|c|c|}
\hline
\textbf{Model}     & \textbf{Image size} & \textbf{Parameter} & \textbf{mAp50} & \textbf{mAp95}  \\ \hline
ResNet            & 800*800            & 20M               & 25,82           & 14,25         \\ \hline
Faster RCNN      & 800*800             & 46M               & 44,75           & 33,74         \\ \hline
ComNet       & 800*800             & 51M               & 53,48           & 30,44          \\ \hline
ViT      & 800*800             & 80M               & 90,33           & 54,94          \\ \hline
Yolov5 & 800*800             & 80M               & 67.59           & 35.62          \\ \hline
SMPNet          & 800*800             & 65M               &    77.31            & 41.25           \\ \hline
DPNetV3               & 800*800             & 127M               & 81,70           & 57,28          \\ \hline
Our model               & 800*800             & 45M               & \textbf{94.57}  & \textbf{60.23}  \\\hline

\end{tabular}

\caption{Comparison results on our datasets}
\label{table:1}
\end{table}
\begin{table}[h!]
\centering
\begin{tabular}{|c|c|c|c|c|}

\hline
\textbf{Model}     & \textbf{Image size} & \textbf{Parameter} & \textbf{mAp50} & \textbf{mAp95}  \\ \hline
ResNet            & 800*800            & 20M               & 21,48           & 11,62         \\ \hline
Faster RCNN      & 800*800             & 46M               & 40.60           & 24,23         \\ \hline
ComNet       & 800*800             & 51M               & 45,67           & 26,86          \\ \hline
ViT      & 800*800             & 80M               & 85,62           & 50,87          \\ \hline
Yolov5 & 800*800             & 80M               & 60.24           & 29.98          \\ \hline
SMPNet            & 800*800             & 65M               &    67.53            & 35.59           \\ \hline
DPNetV3              & 800*800             & 127M               & 82,78           & 56,84          \\ \hline
Our model               & 800*800             & 45M               & \textbf{89.76}  & \textbf{57.42}\\ \hline
\end{tabular}

\caption{Comparison results on public dataset VEDAI}
\label{table:2}
\end{table}

Figure \ref{fig:public_data} shows some results on the VEDAI data, where the first row represents the thermal image, the second row represents the corresponding color image for reference. It is recognizable that our model performs well on common objects like boats, planes, and cars. These images also reflect the challenge with object detection from UAV data when the objects have a very small size. Figures \ref{fig:our_data} display some results on the images collected by our drone under normal and challenging conditions such as urban environments or low light. Here, due to UAV configuration, the thermal data is represented in red color scale instead of greyscale. We also added the third row to provide a close view of our detection results. It can be seen that our model can recognize objects, even those with small sizes and low thermal contrast. Figure \ref{fig:our_data}d shows the clear advantage of using thermal images over color images where the objects can be detected at night.

In another experiment to evaluate the capability of deploying models on embedded devices such as UAVs, we compressed the model through methods that maintain the quality of the model. Specifically, we used transfer learning methods to reduce the model size while preserving its performance. The device used for the experiment was an embedded computer named Jetson AGX 30W. We set up all models to run on the Nvidia device's GPU thread. Table \ref{table:t3} shows the experimental results for different model including the full model, half model, ONNX, TensorRT, and transfer model. It can be seen that with the use of the transfer model, our model can be deployed on the embedded device with real-time speed and high stability. Therefore, the model can be applied to practical systems such as UAVs.

\begin{table}[h!]
\centering
\begin{tabular}{|c|c|c|c|}

\hline
\textbf{Model}     & \textbf{Image size} & \textbf{Speed} & \textbf{Stability}  \\ \hline
Full Model            & 800*800            & 0.02 s     &       1            \\ \hline
Half Model      & 800*800             & 0.005s            &   0.45       \\ \hline
ONNX       & 800*800             & 0.01s        & 0.87            \\ \hline
TensorRT      & 800*800             & \textbf{3ms} & 0.84                       \\ \hline
Transfer model & 800*800             & 4ms & \textbf{0.91}                      \\ \hline

\end{tabular}

\caption{Performance of different models on the Jetson AGX embedded computer}
\label{table:t3}
\end{table}

\section{Conclusion}
In this paper, we have proposed a new neural network model for recognizing objects on thermal images collected by UAVs. Our model uses an improved backbone structure based on YOLOv5 combined with matching classifiers in the neck and prediction head to increase recognition efficiency. Especially, the use of attention blocks before the prediction head greatly increases the ability to identify small objects. Experimental results on both the public and our datasets show that our model performs better than the state-of-the-art methods. It has a low number of parameters and can achieve real-time computation on embedded systems. The model is therefore deployable on real systems such as drones for practical applications.  

\bibliographystyle{ieeetr} 
\bibliography{refs} 

\end{document}